\documentclass[letterpaper, 10 pt, conference]{ieeeconf}  

\IEEEoverridecommandlockouts                              

\usepackage{graphics} 
\usepackage{epsfig} 
\usepackage{mathptmx} 
\usepackage{times} 
\usepackage{amsmath} 
\usepackage{amssymb}  

\usepackage{caption}
\usepackage{subcaption}

\usepackage{tikz}
\usetikzlibrary{shapes.geometric, arrows}

\usepackage{lipsum}

\usepackage{hyperref}
\usepackage{lipsum}
\usepackage{graphicx}
\graphicspath{ {./images/} }
\usepackage{placeins}
\usepackage{diagbox}
\usepackage{xcolor}
\usepackage{authblk}

\newcommand{\vpnote}[1]%
    {\textcolor{cyan}{\textbf{VP: #1}}}
\newcommand{\klnote}[1]%
    {\textcolor{orange}{\textbf{KL: #1}}}
\newcommand{\mlnote}[1]%
    {\textcolor{green}{\textbf{ML: #1}}}

\title{\LARGE \bf
Combining optimal control and learning for autonomous aerial navigation in novel indoor environments}

\author{Kevin Lin, Brian Huo, Megan Hu
\thanks{*Please send email correspondences to kevinlin0@berkeley.edu}}
\affil{University of California, Berkeley}

\begin{document}

\maketitle
\thispagestyle{empty}
\pagestyle{empty}

\begin{abstract}
This report proposes a combined optimal control and perception framework for Micro Aerial Vehicle (MAV) autonomous navigation in novel indoor enclosed environments, relying exclusively on on-board sensor data. We use privileged information from a simulator to generate optimal waypoints in 3D space that our perception system learns to imitate. The trained learning based perception module in turn is able to generate similar obstacle avoiding waypoints from sensor data (RGB + IMU) alone. We demonstrate the efficacy of the framework across novel scenes in the iGibson simulation environment.
\end{abstract}

\section{INTRODUCTION}

Aerial navigation is a well-studied problem in the literature. Researchers have taken advantage of deep learning's ability to implicitly understand visual cues to enable quadrotors to autonomously navigate a variety of environments. However, current solutions to this problem fail to take advantage of aerial robots' ability to move in 3D space. Robots should be able to fly over, under, and around obstacles in 3D space rather than restricting themselves to flying at a prespecified height.

In this paper, we present an algorithm for goal-driven 3D navigation in novel cluttered indoor environments using first-person view images, under the assumption of perfect state estimation. By marrying deep learning and optimal control, we are able achieve good performance in a sample-efficient manner, and generalize to novel environment in simulation with no additional fine tuning.

\section{RELATED WORK}







We base our work off Bansal \textit{et al.}'s LB-WayPtNav \cite{bansal2019-lb-wayptnav}, which demonstrated better performance than end-to-end navigation methods by learning the input to a model-based planning and control module rather than directly learning a control policy. Xia \textit{et al.} used a similar pipeline for control on serial chain manipulators \cite{xia2020relmogen}. We extend this methodology to 3D navigation.

There have been a number of other works that use deep learning for navigation of aerial vehicles through roads \cite{Loquercio_2018}, trails \cite{Smolyanskiy2017Trail}, and indoor environments \cite{Kang2019GtS}, using first person view images. However, none of these other works perform point-based navigation, or allow the drone to change its height. Kauffman \textit{et al.}'s approach is the closest to ours; they learn waypoints to navigate between gates in a 2D drone race track \cite{kaufmann2018deep}. However, they do not consider obstacles or allow for 3D motion.

\begin{figure}[t!]
    \centering
    \includegraphics[ width=\columnwidth]{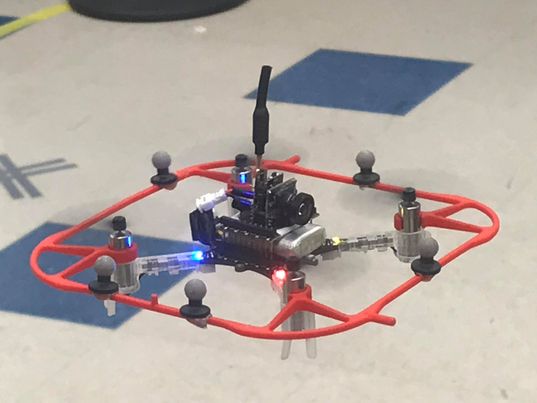}
    \caption{A sample crazyflie 2.0 test platform, with onboard RGB camera and optitrack beacons.}
    \label{fig:aaaaa}
    \vspace{-0.5cm}
\end{figure}

The contributions of this paper are as follows. First, we demonstrate how to leverage existing approaches that combine optimal control and learning based methods for true $3D$ goal-driven quadrotor flight. Second, we demonstrate that the learned policy generalizes to novel environments in simulation. 


\section{Problem Formulation}

We study the problem of autonomous aerial robot navigation in \textit{a priori} unknown indoor environments. We specifically focus on situations where true 3D planning is required. For example, an aerial robot may have to fly over a table, under an overhang, and around a coat rack to reach its destination. The robot's task is to fly to some fixed point in space $p^* = [x^*, y^*, z^*]^T$ without colliding with the environment.

We take our robot to be a quadrotor with a single, forward-facing, monocular RGB camera mounted at a fixed height and pitch from the center of mass. The quadrotor's state can be described by $p = [x, y, z]^T$, the position of the robot's center of mass in the world coordinate frame, $R$ a rotation matrix describing the robot's orientation with respect to this same frame, and their derivatives $\dot{p} = [\dot{x}, \dot{y}, \dot{z}]^T$ and $\dot{R}$. The dynamics of a quadrotor can be found in section II of \cite{lee2010geometric}. We do not consider state estimation in this work, and assume that the robot state, and thus the robot's relative pose to the goal, are perfectly known.

For planning purposes, we use a spline-based trajectory planner, though any other trajectory planner such as minimum snap \cite{kumar2011minsnap} may be used. Our trajectory planner uses a differentially flat representation of the quadrotor model with outputs $p = [x, y, z]^T$ and $\psi = \arctan2(R_{3,2}, R_{2,3})$, the yaw angle of the robot. Given the robot's current state and a waypoint in these coordinates, $w = [w_x, w_y, w_z, w_\psi]^T$, the planner computes the trajectory to the waypoint which minimizes the snap, or fourth derivative, in $x$, $y$, and $z$, as well as the angular acceleration $\ddot{\psi}$ along the trajectory.

In this work, we learn a navigation policy which outputs a waypoint for this min-snap trajectory planner in order to bring the robot to the goal $p^*$ while avoiding obstacles. At a given time $t$, the policy is given an input $(I, x_{rel}, y_{rel}, z_{rel},  \dot{x}_{rel}, \dot{y}_{rel}, \dot{z}_{rel}, \dot{\psi})$ and outpoints a waypoint $w = [w_x, w_y, w_z, w_\psi]^T$. Here, $I$ is the current RGB image from the robot's camera, and $[x_{rel}, y_{rel}, z_{rel}] = R^T(p^* - p)$ and $[\dot{x}_{rel}, \dot{y}_{rel}, \dot{z}_{rel}]^T = R^T \dot{p}$ are the relative position and velocity to the goal, represented \textit{in the body frame} of the robot (not the world coordinate frame). The tasks are performed in novel indoor environments that the robot has never encountered, and whose map or topology are not given to the robot.

\section{Model based learning for 3D point goal visual navigation}

Like the visual navigation technique developed by Bansal \textit{et al.} \cite{bansal2019-lb-wayptnav}, our approach uses three modules for 3D point goal visual navigation: perception, planning and control.

\subsection{Learning based waypoint generation for 3D point goal visual navigation}

\begin{figure*}[tbp!]
    \centering
    \includegraphics[width=2\columnwidth]{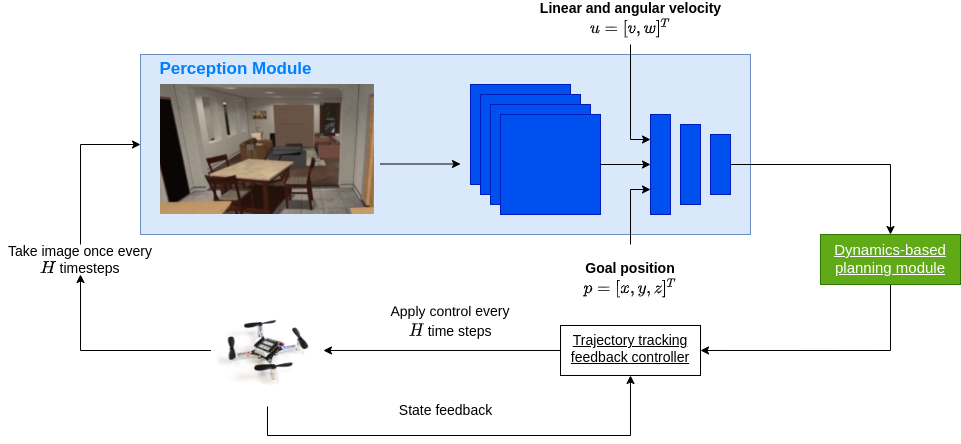}
    \caption{The quadrotor control pipeline relies on a cycle between the perception module, planning module, and flight controller. The perception module takes in RGB images, linear and angular velocity measurements, and a goal position relative to the drone, then uses a neural network to output a waypoint in xyz-yaw coordinates. This waypoint is passed to the planning module, which plots a trajectory to the waypoint. A PID feedback controller tracks the execution of the trajectory as informed by state feedback. Images for the perception module are continuously taken while in flight.}
    \label{fig:Flowchart}
    \vspace{-0.5cm}
\end{figure*}

\subsubsection{Perception} We use a Convolutional Neural Network (CNN) pretrained for a surface normals vision task \cite{sax2018vision} to learn the quadrotor's next waypoint. Specifically, the CNN takes in a $256 \times 256$ pixel RGB image from the onboard camera, the position of the goal in the robot's local coordinate frame and yaw velocity of the robot, and outputs a waypoint $w$. The CNN is trained purely in simulation using an expert policy which uses full knowledge of the world to generate an 'optimal' waypoint.

\subsubsection{Planning} Before passing the waypoint into the planner, we first preprocess it by clipping it into the camera's vision cone. If the waypoint is far outside the cone, such as behind the drone, this generally results in the drone spinning in place. This preprocessing implicitly forces the policy to "look before it leaps" and prevents the drone from colliding with unobserved obstacles.

We represent a trajectory as a vector of time-parameterized continuous functions in the flat coordinates: $\xi(t) = [x(t), y(t), z(t), \psi(t)]$, where $t=0$ represents the current state, and $t=T$ represents the state at trajectory completion, ie the desired waypoint. Given the current drone state $[p_0, \psi, \dot{p}_0, \dot{\psi}_0]$ and a waypoint $[w_x, w_y, w_z, w_\psi]$, the spline-based trajectory planner solves a set of linear equations to output a trajectory in these flat coordinates.

\subsubsection{Control} One benefit of our pipeline is that it's controller-agnostic. As long as the controller is able to reliably track the trajectory planner's, we can use any controller. This also means that our learned policy generalizes well to other (similarly-sized) drones. In our simulated experiments we use a PID controller from Panerati \textit{et al.} \cite{panerati2021learning}. The controller is executed for a time horizon of $H$ seconds (which likely does not lead to completing the trajectory), after which a new image is taken and the process begins again.

\subsection{Data generation and training procedure}

We train the perception module entirely in simulation using a variant of Dataset Aggregation, where the expert action is automatically generated using a combination of model predictive control, $A^*$ planning and full knowledge of the simulation environment. The simulation environment comes from the iGibson \cite{shenigibson} dataset (where the simulation engine is PyBullet \cite{panerati2021learning}), which contains a large variety of 3D scanned environments. To generate training data, we first randomly sample start and goal points for the robot. Then, we let the robot take the expert waypoint with probability $\alpha \in [0, 1]$ and the current CNN output waypoint with probability $1 - \alpha$, where $\alpha$ decays to $0$. We continue until either the robot reaches the goal if until the robot collides. After we collect $X$ new data points, we train the current network on the new dataset $D = D \cup X$. A new single data point consists of the image, goal position, velocities and also the \textit{expert} action. Then, we repeat this process of aggregating more data to the dataset and training the policy on the larger dataset. For each of the scenes in simulation, we collect 20k data points in this manner. \textbf{CNN architecture Details} We use the standard encoder-decoder architecture of a ResNet-50 \cite{Zhang_2018} backbone. For the encoder, we use that of a ResNet-50 pretrained on the surface normals task. We freeze this encoder and only train the decoder part of the network which consists of 5 fully connected layers with ReLU activations. We use an MSE loss and AdamW optimizer with learning rate $10^{-4}$ and weight decay coefficient of $10^{-2}$ on 80k data samples collected using the DAgger procedure described above.

\subsection{Expert Policy}

To generate 'optimal' waypoints for training, we score their resultant trajectories using an MPC-esque cost function which balances distance from obstacles with distance to the goal. Given a prospective waypoint $w$, we compute the corresponding trajectory $\xi(t) = [p(t), \psi(t)]$, and evaluate it using the corresponding cost function:

\begin{align}
    J(w) &= \lambda_1 J_{obs} (\xi) + \lambda_2 J_{dist}(\xi) + \lambda_3 J_{angle} (\xi) \\  
    J_{obs}(\xi) &= -\max (d^{obs}_{cutoff} - d^{obs} (\xi))  \\
    J_{dist}(\xi) &=  d^{goal} (\xi(0))^2 - d^{goal}(\xi(T))^2 \\
    J_{angle}(\xi) &= \frac{1}{T} \sum_{t=0}^T d^{angle} (\xi(t))
\end{align}


The three cost terms respectively represent a penalty for closeness to obstacles, a penalty on the negative progress made to the goal and a penalty on the difference in angle between the `optimal' angle. $d^{obs}$ represents the signed distance to the nearest static obstacle. $d^{goal}(\xi(T))$ represents the minimum collision-free distance between $\xi(T)$ and the goal location. $d^{angle} (\xi(t))$ represents the difference between the angle produced by the trajectory planner at time $t$ and the 'optimal' angle for the trajectory position. We compute the distances to the goal using the Fast Marching Method (FMM) \cite{Sethian1591} and compute 'optimal' angles for points along a trajectory by taking gradients on the distance potentials produced by the FMM map. The $\lambda_i$ terms denote weights for each term. Here, we use $d^{obs}_{cutoff} = 0.4m$, $\lambda_1 = 1, \lambda_2 = 1, \lambda_3 = 0.2$.

We select a waypoint by sampling waypoints and choosing the one with the lowest cost using the MPC cost function outlined above. Since the sample space is large, we increase sample efficiency by biasing our sampling around the output of an $A^*$ planner. However, this planner empirically serves as a good heuristic for sampling.

\section{Experimental Design}

\subsection{Simulation Experiments}

For simulation experiments, we used photo-realistic indoors scenes from the iGibson dataset \cite{shenigibson} and use the CrazyFlie 2.1 model from \textit{gym-pybullet-drones}. We trained the CNN using 4 indoor scenes and test on a 5th held out scene. For the testing scene, we run our policy on 200 randomly sampled test episodes (start, goal position pairs) where the start and goal position pairs involved scenarios like turning a corner, flying above some obstacle(s) and leaving a room. We call a trial successful if the robot reaches reaches within $0.3m$ inside the goal point without collisions. For the simulation experiments, we find that the network is able to successfully generalize to the novel unseen scenes.

\begin{table}[tb] 
\centering
    \caption{Results on simulation experiments} 
    \label{table:1} 
    \begin{tabular}{ | m{1.5cm} | m{1.5cm}| m{1.5cm} | m{1.5cm} |} 
        \hline 
        \textbf{Agent} & \textbf{Expert} & \textbf{LBWayPt3D (Training)} & \textbf{LBWayPt3D (Novel)} \\
        \hline
        \textbf{Success Rate (\%)} & 100 & 93\% & 76\% \\
        \hline
    \end{tabular}
\end{table}

\begin{figure}[!tbp]
  \centering
  \begin{subfigure}[b]{0.5\textwidth}
    \centering
    \includegraphics[width=0.3\textwidth]{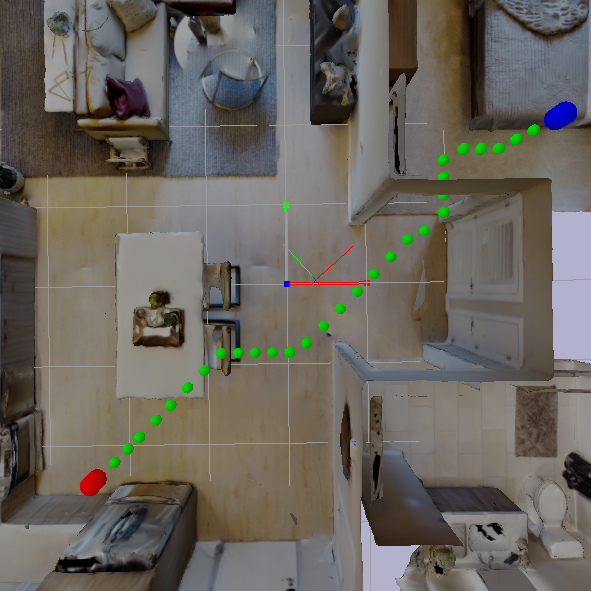}
    \includegraphics[width=0.3\textwidth]{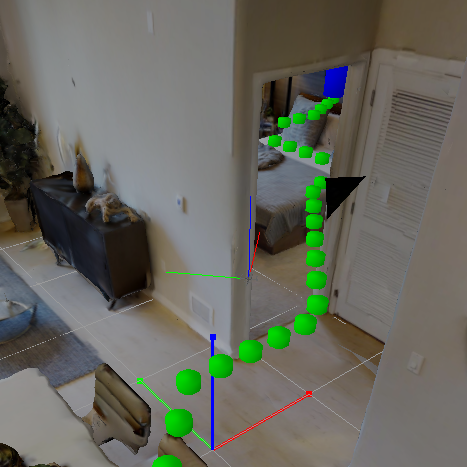}
    \includegraphics[width=0.3\textwidth]{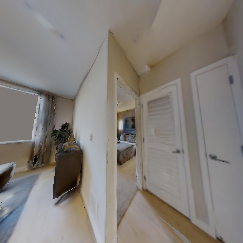}
    \caption{Through doorway}
  \end{subfigure}
  \begin{subfigure}[b]{0.5\textwidth}
    \centering
    \includegraphics[width=0.3\textwidth]{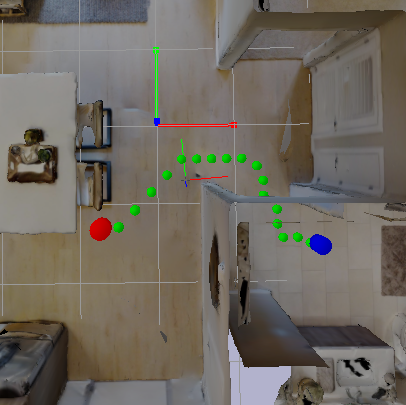}
    \includegraphics[width=0.3\textwidth]{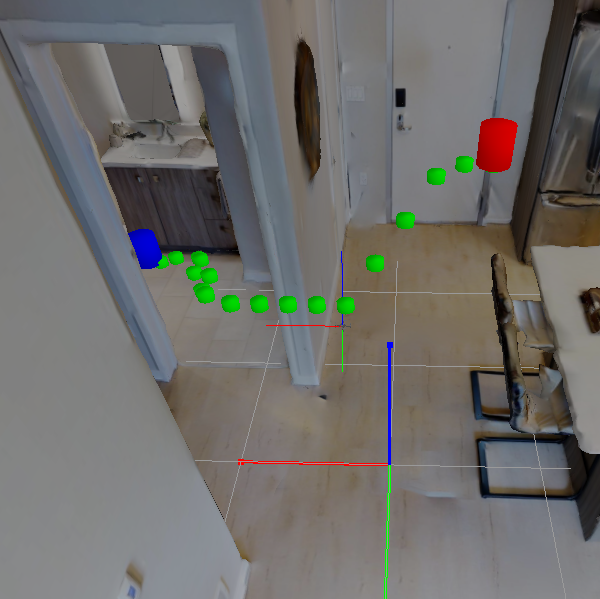}
    \includegraphics[width=0.3\textwidth]{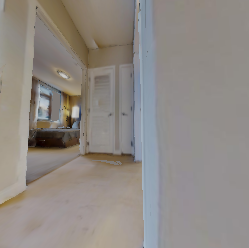}
    \caption{Around corner}
  \end{subfigure}
  \begin{subfigure}[b]{0.5\textwidth}
    \centering
    \includegraphics[width=0.3\textwidth]{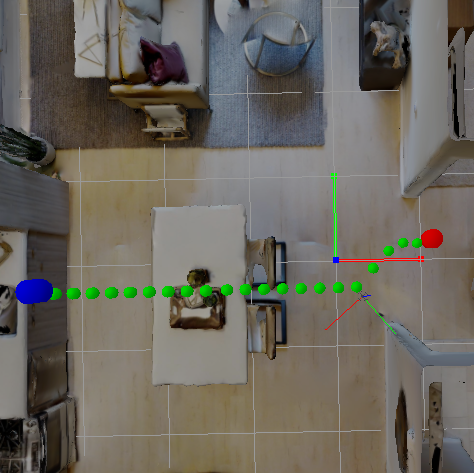}
    \includegraphics[width=0.3\textwidth]{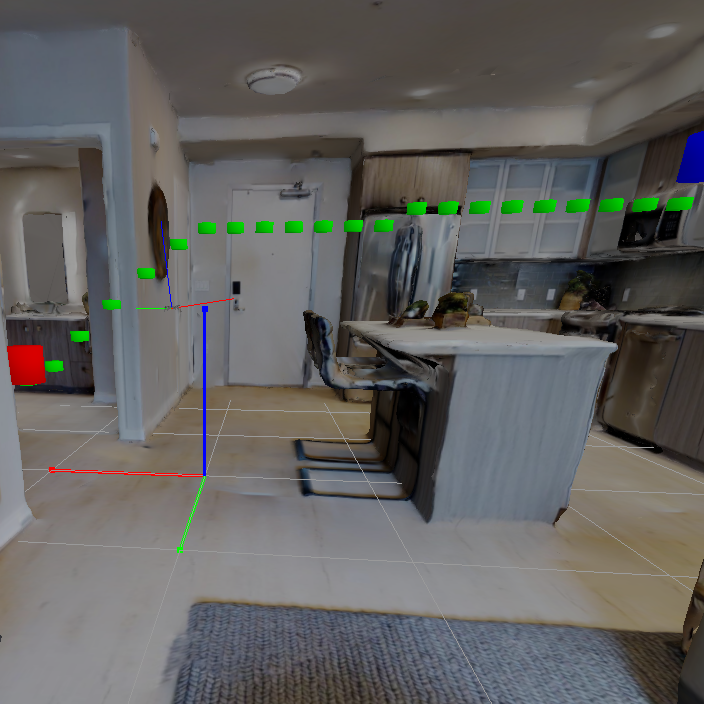}
    \includegraphics[width=0.3\textwidth]{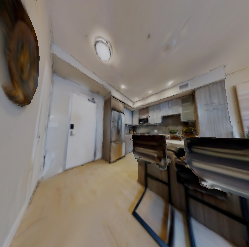}
    \caption{Over table}
  \end{subfigure}
  \caption{Representative simulation experiment episodes on the held out testing scene. Images for each row are given in the order of: bird's eye view, third person view, and robot first person view. The red dot represents the starting point, the blue dot represent the end goal, the green dots represents the path of an $A^*$ planner.}
\end{figure}

Failure Modes. Similar to Bansal \textit{et al.} \cite{bansal2019-lb-wayptnav}, though our policy is able to perform navigation tasks in novel environments, it can only do local reasoning as there is no memory or map implemented in the network. The most prominent failure modes are: a) when the robot is too close to an obstacle, and
b) situations that require ‘backtracking’ from an earlier planned path.

\section{Conclusions and Future Work}

In this report, we present a framework for autonomous aerial goal navigation by combining optimal control techniques and learning methods by building off the LBWayPtNav framework \cite{bansal2019-lb-wayptnav}. Similar to \cite{bansal2019-lb-wayptnav}, we demonstrate, the efficacy of combining optimal control methods with learning methods in simulation. On the other hand, the framework also assumes perfect state estimation and employs a purely reactive policy. These assumptions are not optimal for long range tasks, where incorporating long-term spatial memory is often critical.

\section{Acknowledgements}
We like to thank UC Berkeley's EECS 106B for being the reason we started working on this project, Somil Bansal and the authors of LBWayPtNav for inspiring this particular project, Valmik Prahbu, Somil Bansal and Shankar Sastry for discussions and advice, and David Fridovich-Keil, David McPherson, and Joseph Menke for help with the Crazyflies and Optitrack.

\addtolength{\textheight}{-12cm}   



\section*{APPENDIX}


\bibliographystyle{IEEEtranS.bst}
\bibliography{bibtex}

\end{document}